\documentclass{article}

\usepackage[square,numbers]{natbib}
\usepackage[preprint]{neurips_distshift_2022}

\usepackage{xspace}
\usepackage[utf8]{inputenc} % allow utf-8 input
\usepackage[T1]{fontenc}    % use 8-bit T1 fonts
\usepackage[colorlinks]{hyperref}       % hyperlinks
\usepackage{url}            % simple URL typesetting
\usepackage{booktabs}       % professional-quality tables
\usepackage{amsfonts}       % blackboard math symbols
\usepackage{amsmath}

\DeclareMathOperator*{\argmin}{arg\,min}
\DeclareMathOperator*{\probe}{\text{probe}}
\usepackage{nicefrac}       % compact symbols for 1/2, etc.
\usepackage{microtype}      % microtypography
\usepackage[table]{xcolor}
\usepackage{graphicx, graphics}
\usepackage{color}
\usepackage{subcaption}
\usepackage{amsthm}
\usepackage[english]{babel}
\usepackage[T1]{fontenc}
\usepackage{amsmath}
\usepackage{amssymb}
\usepackage{amsfonts}
\usepackage{multirow}
\usepackage{mathtools}
\usepackage{breqn}
\usepackage{bbm}
\usepackage{dsfont}
\usepackage{graphicx}
\usepackage{natbib}
\usepackage{cases}
\usepackage[colorinlistoftodos]{todonotes}
\usepackage{booktabs}
\usepackage{array}

\renewcommand{\hat}{\widehat}

\newtheorem*{remark*}{Remark}

\newtheorem*{observation*}{Observation}

\numberwithin{equation}{section}

\newcommand{\cT}{\mathcal{T}}

\usepackage{bbm}
\newcommand{\naive}{SGD\xspace}
\definecolor{lgray}{RGB}{245,245,245}
\newcommand{\commentout}[1]{}

\title{Improving Representational Continuity with Supervised Continued Pretraining}

\author{%
  Michael Sun \\
  Department of Computer Science\\
Stanford University\\
  \texttt{msun415@cs.stanford.edu} \\
  \And
  Ananya Kumar \\
  Department of Computer Science \\
  Stanford University\\
  \texttt{ananya@cs.stanford.edu} \\
  \AND
  Divyam Madaan \\
  New York University \\
  \texttt{divyam.madaan@nyu.edu,} \\
  \And
  Percy Liang \\
  Stanford University \\
  \texttt{pliang@cs.stanford.edu} \\
}

\def\shownotes{1}
\ifnum\shownotes=1
\newcommand{\authnote}[2]{[#1: #2]}
\else
\newcommand{\authnote}[2]{}
\fi

\begin{document}

\maketitle

\begin{abstract}We consider the continual representation learning setting: sequentially pretrain a model $M'$ on tasks $\cT_1, \ldots, \cT_T$, and then adapt $M'$ on a small amount of data from each task $T_i$ to check if it has forgotten information from old tasks. Under a kNN adaptation protocol, prior work shows that continual learning methods improve forgetting over naive training (\naive). In reality, practitioners do not use kNN classifiers---they use the adaptation method that works best (e.g., fine-tuning)---here, we find that strong continual learning baselines do worse than naive training. Interestingly, we find that a method from the transfer learning community (LP-FT) outperforms naive training and the other continual learning methods. Even with standard kNN evaluation protocols, LP-FT performs comparably with strong continual learning methods (while being simpler and requiring less memory) on three standard benchmarks: sequential CIFAR-10, CIFAR-100, and TinyImageNet. LP-FT also reduces forgetting in a real world satellite remote sensing dataset (FMoW), and a variant of LP-FT gets state-of-the-art accuracies on an NLP continual learning benchmark.
\end{abstract}

\section{Introduction}
\label{sec:intro}

% Introduce the continued pretraining task
We consider the setting of continual representation learning~\cite{rao2019continual, madaan2022representational}.
% where we first sequentially pretrain a model $M'$ on a sequence of tasks, and then fine-tune $M'$ on a task where we have a small amount of labeled data.
As a motivating example, suppose we are developing a foundation model~\cite{bommasani2021opportunities} for satellite remote sensing~\cite{jean2016combining}.
% that can classify a satellite image into a category such as ``hospital", ``impoverished settlement", etc.
A resource rich organization \emph{pretrains} a model $M$ on lots of satellite data from North America.
Over time, the organization needs to upgrade the capabilities of its model by incorporating more geographical regions---retraining the model is expensive so it takes $M$ and \emph{continues the pretraining process} on data from Africa to get a model $M'$.
% The organization then releases the model $M'$. 

% Introduce the evaluation task
The organization releases the pretrained model, and resource constrained academic labs across the world \emph{use the updated model $M'$} for important applications---a common way to use the model $M'$ is to \emph{fine-tune} it on a small amount of task-specific labeled data.
We want the updated model $M'$ to be useful for both new data (from Africa) and old data (from North America).

% In continual learning, there are continuum of $T$ tasks $T_1,\ldots, T_T, 1 \le t \le T$. For each task $t$, we have a training dataset $\mathcal{D}_{tr}^{(t)}$, few-shot dataset $\mathcal{D}_{ft}^{(t)}$, evaluation dataset $\mathcal{D}_{te}^{(t)}$ sampled from $X_t \times Y_t$.
% % \subsection{Models and losses}
% We parameterize the models by a base feature extractor $\theta \in \mathcal{B}$ and task-specific head $\phi_i \in \mathcal{V}$, where we predict $h_{\phi_i}( f_{\theta}(x) )$ on task $t$. $f_{\theta}(x) \in R^k$ are lower-dimensional features corresponding to the high-dimensional input $x$.
% $f_{\theta}$ is then a neural network or a linear projection to a lower dimensional feature space, while $h_{\phi}$ is typically a linear head. For a continual learner, we the learning objective can be defined as:
% $$\mathcal{L}_t(\phi, \theta) = \sum_{x, y \in D_{tr}^{(t)}} \ell(h_{\phi}(f_{\theta}(x)), y),$$

% Formal setting.
More formally, we sequentially pretrain a model on a sequence of tasks $\cT_1,\ldots, \cT_T$ to get a model $f_{\theta_T}$---that is, we pretrain on data from $\cT_1$, then data from $\cT_2$, etc.
Naively pretraining in this way (which we call \naive) leads to \emph{forgetting}---the model performs poorly on data from older tasks such as $\cT_1$.
Prior works in continual learning propose a variety of methods to reduce forgetting: popular methods include Synaptic intelligence (SI), Dark Experience Replay (DER), and Unsupervised Continual Learning (UCL).
To examine if $f_{\theta_T}$ has retained knowledge about every task $\cT_i$, we \emph{adapt} $f_{\theta_T}$ using a small amount of labeled data from task $\cT_i$, and evaluate its accuracy on task $\cT_i$.

% There are many ways to probe.
Practitioners will typically use the adaptation method that gets the highest accuracy for their use case—in this setting, we find that \emph{continual learning methods such as UCL, DER, and SI, can actually do worse than naive training (\naive).}
For each task, we consider adapting the pretrained model with three adaptation methods (training a kNN or linear classifier on frozen representations, or full fine-tuning of the model) on a small amount of task specific data.
Prior work uses the kNN evaluation protocol---partly because this is cheap to evaluate.
Under this scheme, we indeed see that continual learning methods improve over naive training.
However, when considering the best adaptation method, we find that strong continual learning baselines (SI, DER, and UCL) can perform even worse than naive training---see Table \ref{tab:fewshot-probes-eval}.
For example, under a kNN evaluation protocol, SI gets 86\% which is 8\% better than naive \naive (78\%), but under the best adaptation method SI gets 91\% accuracy which is \emph{2\% lower than naive \naive} (93\%).

% LP-FT
However, we find that a method from the transfer learning community \emph{(LP-FT) \cite{LPFT} outperforms naive \naive}.
For each new task $\cT_i$, naive \naive updates the entire model via gradient descent---this can distort representations learned from previous tasks.
Instead,\cite{LPFT} proposes first training the linear `head' layer on the new task $\cT_i$ (to find the best way to use existing representations for the new task) and then fine-tune the entire model to incorporate information from $\cT_i$ into its representations.
LP-FT consistently improves over naive \naive under all adaptation methods, and gets the highest overall accuracy.
Even under the exact kNN evaluation protocol used in \cite{madaan2022representational}, LP-FT matches or outperforms continual learning methods on three popular datasets: Sequential-CIFAR10 \cite{krizhevsky2012imagenet}, Sequential-CIFAR100 \cite{krizhevsky2012imagenet}, and Sequential-TinyImageNet \cite{deng2009imagenet}.

% LP-FT is general
Finally, we find that LP-FT performs well in two other domains as well.
1. We consider continual representation learning in a real world satellite remote sensing task (Functional Map of the World~\cite{christie2018fmow})---LP-FT (56\%) outperforms naive \naive (53\%). 
2. LP-FT gets \emph{state-of-the-art accuracies on an NLP continual learning setting}~\cite{Ke2021AdaptingBF} where they update BERT for a sequence of sentiment analysis tasks.~\cite{Ke2021AdaptingBF} propose a new continual learning method that outperforms 12 strong baselines. We show that a variant of LP-FT gives a further 2\% accuracy boost over their strongest method.

% \subsection{Paper ID}
% Make sure that the Paper ID from the submission system is visible in the version submitted for review (replacing the ``*****'' you see in this document).
% If you are using the \LaTeX\ template, \textbf{make sure to update paper ID in the appropriate place in the tex file}.
\section{Related Work}
\label{sec:related-work}
Over the years, many continual learning approaches have been proposed in three main categories.  \emph{Regularization}-based methods~\cite{zenke17si, lopez2017gradient, chaudhry2018efficient} minimize the drift of model representations during sequential learning to prevent the forgetting of learned knowledge. \emph{Architectural}-methods~\cite{rusu2016progressive, YoonJ2018iclr, yoon2020apd} expand the network structure to learn the task-independent and task-shared knowledge across the sequence of tasks. \emph{Replay}-based methods~\cite{sinha2020experience, Aljundi2019GradientBS, buzzega2020dark} select and revisit a representative subset of the past-task examples during the training of future tasks to alleviate forgetting. While most of these methods were restricted to supervised settings, recent works~\cite{madaan2022representational, fini2022self} have extended them to continual learning with the unlabelled data stream.

\section{Preliminaries}
\label{sec:formatting}

In continual learning, there is a continuum of $T$ tasks $\mathcal{T}_1,\ldots, \mathcal{T}_T$. For each task $\mathcal{T}_t$, we have a training dataset $\mathcal{D}_{tr}^{(t)}$, few-shot dataset $\mathcal{D}_{ft}^{(t)}$, evaluation dataset $\mathcal{D}_{te}^{(t)}$ sampled from a distribution $P_t$ over $\mathcal{X}_t \times \mathcal{Y}_t$.
% \subsection{Models and losses}
We parameterize the models by a base feature extractor $\theta \in \mathcal{B}$ and task-specific head $\phi_i \in \mathcal{V}$ to predict $h_{\phi_t}( f_{\theta}(x) )$ on task $t$, where $f_{\theta}(x) \in R^k$ maps inputs into a lower dimensional feature space, and $h_{\phi}$ represents a linear head. 

%-------------------------------------------------------------------------
\subsection{Training}

Given a loss function $\ell : X \times Y \rightarrow R_{\geq 0}$ (for example, the cross-entropy loss), the loss on task $t$ is the average of the loss $l$ over the training set for task $i$:
\begin{equation}\mathcal{L}_t(\phi, \theta) = \sum_{x, y \in D_{tr}^{(t)}} \ell(h_{\phi}(f_{\theta}(x)), y)\end{equation}

\subsection{Linear probe finetuning for CL}
% The usual convention is to pretrain $\theta$ on lots of data, then transfer learning to learn $\phi$. In continual learning, $\theta$ and $\phi$ are then adapted on a set of tasks, and we do not use pretraining to adapt $\theta$ and $\phi$ from scratch.
% There are many adaptation methods, but two widely used methods are linear probing and fine-tuning. In continual learning, a new head is added for every task. Linear-probing only adapts the head $\phi_i$, while fine-tuning adapts both the feature extractor $\theta$ and the head $\phi_i$ using gradient descent on the training data ${P_i}_{train}$ for any task.
% We can finetune both $\phi_i$ and shared representation $f_{\theta}$ for each task $i$, which is good ID, but because $\phi_i$ is randomly initialized, it can be unaligned to $f_{\theta}$ \cite{LPFT}.
% If the head is not aligned, it will have to change a lot throughout the trajectory of fine-tuning a new task, and shown to perform worse OOD - the change to the random head and good feature extractor are coupled together, causing feature distortion. 
% If the head is already aligned at the start to the new task, the feature extractor has to only modify slightly, and is mostly preserved. In context of CL, it is critical the feature extractor is preserved, and most successful CL algorithms like regularization impose ways to prevent it from changing. LPFT can do same thing but without incurring additional cost like requiring past data, weights, or extra memory compared to other “oracle” baselines we compare with.

We start with $\hat{\theta_0}$ initialized randomly. For each task $1\le t \le T$, LP-FT first trains the head $\phi$, and then jointly optimizes the entire model $(\phi, \theta)$.
\begin{align}
    \hat{\phi_{t,lp}} &= \argmin_{\phi} L_t(\phi, \hat{\theta}_{t-1}) \\
    \hat{\phi_{t}}, \hat{\theta_t} &= \argmin_{\phi, \theta} L_t(\phi, \theta) \mbox{, initialized at }\phi = \hat{\phi_{t,lp}},  \theta = \hat{\theta}_{t-1},
\end{align}
where we approximate the $\argmin$ using stochastic gradient descent. We apply LP-FT on both a standard ResNet18 classifier, and the B-CL architecture from \cite{Ke2021AdaptingBF} and show its efficacy in both vision and NLP. More architecture-level decisions can be found in Appendix \ref{app:app-2}. 
% \ak{need to describe the other methods}

% Intuitively, this works because continual learning can be regarded as a sequence of distribution shifts (the classes are different) with the important assumption that there exists a good feature extractor that is shared for all the tasks. If the head is randomly initialized, it will have to change a lot throughout the trajectory of fine-tuning a new task. As a result, the change to the random head and good feature extractor are coupled together, causing feature distortion. If the head is already aligned at the start to the new task, the feature extractor has to only modify slightly, and is mostly preserved.

%-------------------------------------------------------------------------
% \subsection{Evaluation protocols}
\subsection{Evaluation}\label{sec:evaluation}
At evaluation time, we consider the model $\theta_T$ at the end of continual learning. Our goal is to evaluate how good the quality of representations in $\theta_T$ are for older tasks by evaluating $\theta_T$ on (a typically small amount of) data from each task $i$. We probe the model on each of the tasks $i$:
\begin{equation}
    \hat{\xi_{i}}, \hat{\theta_i} = \probe(\theta_T, D_{ft}^{(i)})
\end{equation}
We then get the accuracy $A_i$ on task $i$, and finally measure the average $A$ over all the tasks:
\begin{equation}
    A = \frac{1}{T}  \sum_{i=1}^T A_i \mbox{, where } A_i = \frac{1}{| D_{{te}}^{(i)} |} \sum_{x. y \in D_{te}^{(i)}} \mathbbm{1}[h_{\hat{\xi_{i}}}(f_{\hat{\theta_i}}(x)) = y].
\end{equation}
We investigate three different options for $\probe$: 1. Train a kNN classifier on frozen representations produced by the model $f_{\hat{\theta_T}}$, 2. Train a linear probe on the model representations, 3. Fine-tune the entire model parameters via LP-FT. 

\section{Probe Definitions}
% \label{app:app-4}

We include formal definitions for the three different instantiations  of the probing classifier used for our evaluation.

\begin{enumerate}
\item {\bf Linear probe} 

We train a linear classifier on frozen representations produced by the model $\theta_T$:
\begin{equation}
\begin{split}
    \probe(\theta_T, D_{ft}^{(i)}) :&= \argmin_{\xi} L_i(\xi), \theta_T \mbox{, where } \\ 
    L_i(\xi)     &=  \sum_{x, y \in D_{ft}^{(i)}} l(h_{\xi}(f_{\theta_T}(x)), y)
    \end{split}
\end{equation}
\item  {\bf KNN probe}
$\xi$ follows \cite{wu18knn} by building a nearest-neighbor classifier based on ${\theta_T(x) : x \in D_{ft}^{(i)}}.$ 
% \ak{not sure what this means}
\item {\bf LPFT probe}
We define a loss function:
$$L_i(\xi, \theta) = \sum_{x, y \in D_{ft}^{(i)}} l(h_{\xi}(f_{\theta}(x)), y).$$
For each task $i$, we first train the head $\xi$ and then fine-tune the entire model $\xi, \theta_T$.
\begin{align}
    \hat{\xi_{i}}_{lp} &= \argmin_{\xi} L_i(\xi, \theta_{T}) \\
    \probe(\theta_T, D_{ft}^{(i)}) :&= \argmin_{\xi, \theta} L_i(\xi, \theta)
\end{align}
initialized at $\xi = \hat{\xi_{i}}_{lp}, \theta = \theta_T.$
% \ak{Michael, can you move the definitions below to the appendix and reformat, will save space}
% \ak{Replace double dollar with the equation environment, so it's numbered}
\end{enumerate}

% Do not change any aspects of the formatting parameters in the style files.  In
% particular, do not modify the width or length of the rectangle the text should
% fit into, and do not change font sizes (except perhaps in the
% \textbf{References} section; see below). Please note that pages should be
% numbered.

\section{Experiments}
We show quantitative results on popular benchmarks Sequential-CIFAR-10 $(T=5)$, CIFAR-100 $(T=20$), Tiny-ImageNet $(T=10)$, as well as a benchmark of 19 Aspect Sentiment Classification Datasets \cite{Ke2021AdaptingBF} $(T=19)$ and an initial result on real-world satellite data Functional Map of the World $(T=6)$. We abbreviate these as C10, C100, Tiny, ASC, and FMOW. We show that LPFT significantly improves over fine-tuning on C10, C100, and Tiny, and is almost as good as the best unsupervised continual learning algorithms and SOTA supervised continual learning algorithms. In order to better evaluate representations, we introduce an evaluation protocol where we are given a small percentage of data to tune on at test time and show that the benefit from popular regularization and replay continual learning algorithms is largely negated when we follow this evaluation protocol.

\subsection{Experimental Setup}
At training time, we tune every supervised continual learning method by averaging over 3 random seeds and sweeping over 6 learning rates. Meanwhile, we use the best method-specific hyperparameters from \cite{buzzega2020dark} and \cite{madaan2022representational} for DER and SI, adding group normalization which was found to work better across all supervised methods. We use the best hyperparameters from \cite{madaan2022representational} directly for UCL, reproducing their best runs. For more details, see section \ref{sec:app-1}.

To prepare the best runs for probe evaluation, we select the hyperparameters corresponding to the best confidence interval, using the KNN probe accuracy as the validation and test metric. Then, we rerun 3 random seeds with those hyperparameters and obtain 3 corresponding checkpoints with the highest average test accuracy in the span of the final task. For KNN probe, we follow the same hyperparameters as \cite{madaan2022representational}. For linear probes, we using lbfgs solver to sweep over 100 regularization values, and evaluate accuracy using the best classifier. For LPFT probe, we sweep over the same 6 learning rates as training and train for the same number of epochs as during the training.

\subsection{Ranking of Methods Depend on Evaluation Protocol}
First, we compare the methods using the original evaluation protocol, which evaluates the mean test accuracy using the KNN classifier over all tasks at the end of training. The results from \ref{tab:full-knn-eval} are shown.

Second, we apply our evaluation protocol as described in \ref{sec:evaluation}, which uses 10\% of the data as fewshot to train various probes. As shown in \ref{tab:fewshot-probes-eval}, SI and DER are the best two methods with the standard KNN and LP probes, but \naive and LPFT are the best two methods with the LPFT probe. With the exception of UCL, all methods obtain a better accuracy with LPFT probe, suggesting it is the probe that practitioners will elect to use to maximize accuracy. \naive and LPFT especially yield high-performing probes that beat DER.

\begin{table}[h!]\centering
\caption{LPFT does better than UCL and DER and almost as well as SI. On Tiny, LP-FT does better than all other methods---6\% higher accuracy than the next best method.}
% \begin{tabular}{lll}
% \toprule
%      & \multicolumn{2}{l}{KNN probe}           \\
%      \midrule
% FT   & \multicolumn{1}{l}{5. 88.92} & 5. 71.27 \\ 
% LPFT & \multicolumn{1}{l}{2. 91.65} & 2. 79.24 \\
% UCL  & \multicolumn{1}{l}{4. 90.11} & 4. 75.42 \\ 
% \midrule
% DER  & \multicolumn{1}{l}{3. 90.38} & 3. 76.97 \\ 
% SI   & \multicolumn{1}{l}{1. 92.73} & 1. 79.63 \\ 

% \bottomrule
% \end{tabular}

\begin{tabular}{lllll}
\hline
\multicolumn{1}{c}{}  & \multicolumn{4}{l}{KNN probe}                                                \\ \midrule
                      & C10               & C100              & Tiny              & Average           \\ \midrule
\naive & 5. 88.92          & 5. 71.27          & 4. 57.12          & 5. 72.44          \\
LPFT                  & 2. 91.65          & 2. 79.24          & \textbf{1. 64.25} & \textbf{1. 78.38} \\
UCL                   & 4. 90.11          & 4. 75.42          & 3. 58.31          & 4. 74.61          \\ \midrule
DER                   & 3. 90.38          & 3. 76.97          & 5. 56.97          & 3. 74.77          \\
SI                    & \textbf{1. 92.73} & \textbf{1. 79.63} & 2. 58.55          & 2. 76.97          \\ \bottomrule
\end{tabular}
\label{tab:full-knn-eval}
\end{table}

\begin{table}[h!]\centering
\caption{We find that using the best probe trained on fewshot data, continual learning methods do worse than \naive, while LPFT does better. We tested on 10\% few-shot data under our various probes’ evaluation schemes. The average metric over C10 and C100 is reported. The full table can be found in \ref{app:app-2}. }

\begin{tabular}{lllll}
\toprule
        & \multicolumn{1}{c}{KNN probe} & \multicolumn{1}{c}{LP probe} & \multicolumn{1}{c}{LPFT probe} & \multicolumn{1}{c}{Best probe} \\ \midrule
\naive     & 5. 77.64                         & 4. 88.30                        & 2. 93.46                          & 2. 93.46                          \\
UCL  & 4. 82.45                         & 5. 88.21                        & 5. 87.06                          & 5. 88.21                          \\
LPFT & 3. 83.95                         & 3. 90.20                        & \textbf{1. 93.74}                 & \textbf{1. 93.74}                 \\ \midrule
SI      & \textbf{1. 85.87}                & \textbf{1. 90.91}               & 4. 91.05                          & 4. 91.05                          \\
DER     & 2. 84.78                         & 2. 90.68                        & 3. 93.26                          & 3. 93.26                          \\ 
\bottomrule
\end{tabular}

\label{tab:fewshot-probes-eval}
\end{table}

\subsection{LPFT Maintains Strong Performance and Further Gains from Proper Evaluation}
From Table \ref{tab:full-knn-eval}, we see LPFT at training time significantly improves over \naive (2.73\% on C10, 7.97\% on C100) with the KNN evaluation protocol in \cite{madaan2022representational}, outperforming DER and second only to SI. This simple and effective technique refutes prior conclusions that supervised finetuning cannot outperform unsupervised learning unless replay or regularization is applied. LPFT is also not data-intensive. From Table \ref{tab:fewshot-probes-eval}, we see with only 10\% fewshot data, applying LPFT probe to \naive and LPFT yields linear probes that beats DER (on C100), or comes within 0.28\% of it (C10). Meanwhile, UCL does not gain from LPFT at few-shot evaluation time, losing 2.46\% and gaining 0.19\% from a linear probe (-1.15\% on average), implying its representation quality hinges on having access to the full dataset. Using additional data at test time helps “simple” methods defeat their more involved counterparts, without incurring additional cost like memory from replay buffer (DER) or past weights for information accumulation (SI). We highlight this by partitioning the table into methods that incur extra memory (bottom) and methods that don't (top).

\subsection{LPFT Couples with SOTA CL Methods for Consistent Gains}
In this section, we also include results coupling LPFT with SI and DER in Table \ref{tab:ablations}. We see that for every supervised continual learning method - finetuning, SI, DER - adding LPFT either results in gains that outperform UCL or comes near confidence interval of it. These experiments challenge the prior belief that unsupervised methods are becoming the SOTA way to continually learn representations \cite{madaan2022representational}. 

\label{sec:app-1}
\begin{table}[h!]\centering
\caption{We report all supervised methods with batch norm (BN) and with group normalization (GN). Methods with +SK mean sklearn is used to obtain the linear probe.}

\begin{tabular}{lrr}
\toprule
                                   & \textsc{C10}        & \textsc{C100}      \\ \midrule
FT (BN)                    & 88.19 ($\pm$1.02)  & 70.89 ($\pm$1.92) \\
FT (GN)                    & 88.92 ($\pm$0.31)  & 71.27 ($\pm$1.23) \\
LPFT (BN)                 & 89.73 ($\pm$0.48)  & 75.38 ($\pm$1.66) \\
LPFT (GN)                 & 91.18 ($\pm$0.21)  & 76.42 ($\pm$1.38) \\
LPFT + SK (GN) & \textbf{91.65 ($\pm$0.05)}  & \textbf{79.24 ($\pm$0.52)} \\

UCL (original, BN)  & 90.11 ($\pm$0.12)  & 75.42 ($\pm$0.78) \\
\midrule
DER (ours, GN)             & 90.38 ($\pm$0.53)  & 76.97          \\
DER + LPFT (GN)           & 91.07 ($\pm$0.25)  & 78.21          \\
DER + LPFT + SK (GN)   & \textbf{91.65 ($\pm$0.32)}  & \textbf{80.56 ($\pm$0.15)} \\
UCL + DER (original, BN)  & 91.22 ($\pm$0.30)  & 77.27 ($\pm$0.30) \\
\midrule
SI (BN)                    & 91.02 ($\pm$0.40)  & 78.85 ($\pm$0.88) \\
SI (GN)                    & \textbf{92.73 ($\pm$0.07)} & \textbf{79.63 ($\pm$0.10)} \\
SI + LPFT (GN)            & 92.26 ($\pm$0.15)  & \textbf{80.01 ($\pm$0.32)} \\
SI + LPFT + SK (GN)  & 92.64 ($\pm$0.07)  & 78.03 ($\pm$0.63) \\ 
UCL + SI (original, BN)  & \textbf{92.75 ($\pm$0.06)}  & \textbf{80.08 ($\pm$1.30)} \\
\bottomrule
\end{tabular}

\label{tab:ablations}
\end{table}
For both the LPFT method and LPFT probe evaluation, our ablation study considers two ways of probing - either training the linear head with \naive for 25 epochs with the rest of the model frozen, or using sklearn (+sklearn) to obtain the probe via the lbfgs logistic regression solver from sklearn. For LPFT training, both ways are followed by 25 epochs of finetuning. We find sklearn’s logistic regression obtains a better linear probe than training by sweeping over 100 regularization values within the logspace of $[10^{-7}, 10^2]$, and that using the obtained weights to set the probe improves the performance of LPFT and LPFT-coupled methods. For the results in \ref{tab:full-knn-eval}, we performed all instances of linear probing with sklearn's probe. 

We tuned our supervised methods for the best learning rate over $[0.003,0.01,0.03,0.1,0.3,1.0]*\text{batch size}/256$, with the adjustment rule from \cite{madaan2022representational}. For UCL methods, we did not tune the learning rate due to our computational constraints: UCL requires 200 epochs (instead of 50) to converge, and employs batch size of $256$ instead of $32$ used for supervised methods. Instead, these hyperparameters were tuned extensively by \cite{madaan2022representational}, so we use them as is.

\subsection{LPFT Generalizes to Other Domains and Scales to Larger Datasets}
\begin{table}[!htp]\centering
\caption{(left) LP-FT gives a further boost of 0.55\% and 1.41 Macro F1 compared to B-CL \cite{Ke2021AdaptingBF}, the state-of-the-art method on an aspect sentiment classification CL benchmark. (right) LP-FT also improves over naive training on a real-world satellite remote sensing dataset (FMoW) where 6 tasks correspond to 6 continents, getting 3\% higher average accuracy.}

\begin{subtable}{0.49\columnwidth}\centering
\caption{}
\label{tab:b-cl}
\begin{tabular}{lrr}
\toprule
            & Accuracy       & Macro F1       \\ 
            \midrule
B-CL        & 89.51 ($\pm$0.55) & 82.05 ($\pm$1.30) \\ 
B-CL + LPFT & 90.06 ($\pm$0.77)  & \textbf{83.46} ($\pm$0.67) \\ 
\bottomrule

\end{tabular}
\end{subtable}
\begin{subtable}{0.49\columnwidth}\centering
\caption{}
\label{tab:fmow}
\begin{tabular}{lll}

\toprule
     & KNN probe \\ 
     \midrule
\naive   & 53.22         \\
LPFT & 56.03       \\
\bottomrule
\end{tabular}

\label{tab:fmow}
\end{subtable}

\end{table}
In Table \ref{tab:b-cl}, we see that $\phi$ and $\xi$ can be instantiated as task-specific parameters for a general task-incremental CL framework in the domain of NLP, and LPFT can still achieve gains. We follow their setup and report the average over 5 random sequences of the 19 tasks. 

We also apply LPFT to FMOW, a real-world dataset of satellite imagery, and the initial result shows LPFT can bring immediate gains even on a messy real-world dataset where the most resourceful region has >400x samples as the least resourceful region.

LPFT's numbers are generally more competitive for larger datasets. As detailed in Appendix Table \ref{tab:fewshot-probes-eval-full}, the gap between LPFT and \naive increases from 3.94\% (KNN probe), 1.01\% (linear probe) to 8.69\% and 2.7\% respectively on C100 compared to C10. On Tiny, we see LPFT to be the dominant method, achieving 7.28\% and 1.67\% on top of the second best methods under KNN (DER) and linear probe (SI) evaluations respectively. 

%------------------------------------------------------------------------

\section{Discussion and Conclusion}

This works aims to learn better representations via continued pretraining and find better protocols to evaluate them. We introduce a probe evaluation framework that changes the ranking of continual learning methods. We introduce a simple yet effective technique that boosts performance across all datasets, can be applied in different domains and settings, couples with existing continual learning methods, and does not incur additional costs. 
% In this work, we focus on the supervised representation learning setting in continual learning, but do not consider other settings like class-incremental or task-agnostic. For our follow-up work, we will also include class-il and task-il numbers.

%%%%%%%%% REFERENCES
{\small
\bibliography{references}

\begin{thebibliography}{21}
\providecommand{\natexlab}[1]{#1}
\providecommand{\url}[1]{\texttt{#1}}
\expandafter\ifx\csname urlstyle\endcsname\relax
  \providecommand{\doi}[1]{doi: #1}\else
  \providecommand{\doi}{doi: \begingroup \urlstyle{rm}\Url}\fi

\bibitem[Aljundi et~al.(2019)Aljundi, Lin, Goujaud, and
  Bengio]{Aljundi2019GradientBS}
R.~Aljundi, M.~Lin, B.~Goujaud, and Y.~Bengio.
\newblock Gradient based sample selection for online continual learning.
\newblock In \emph{Advances in Neural Information Processing Systems
  (NeurIPS)}, 2019.

\bibitem[Bommasani et~al.(2021)Bommasani, Hudson, Adeli, Altman, Arora, von
  Arx, Bernstein, Bohg, Bosselut, Brunskill, Brynjolfsson, Buch, Card,
  Castellon, Chatterji, Chen, Creel, Davis, Demszky, Donahue, Doumbouya,
  Durmus, Ermon, Etchemendy, Ethayarajh, Fei-Fei, Finn, Gale, Gillespie, Goel,
  Goodman, Grossman, Guha, Hashimoto, Henderson, Hewitt, Ho, Hong, Hsu, Huang,
  Icard, Jain, Jurafsky, Kalluri, Karamcheti, Keeling, Khani, Khattab, Koh,
  Krass, Krishna, Kuditipudi, Kumar, Ladhak, Lee, Lee, Leskovec, Levent, Li,
  Li, Ma, Malik, Manning, Mirchandani, Mitchell, Munyikwa, Nair, Narayan,
  Narayanan, Newman, Nie, Niebles, Nilforoshan, Nyarko, Ogut, Orr,
  Papadimitriou, Park, Piech, Portelance, Potts, Raghunathan, Reich, Ren, Rong,
  Roohani, Ruiz, Ryan, Ré, Sadigh, Sagawa, Santhanam, Shih, Srinivasan,
  Tamkin, Taori, Thomas, Tramèr, Wang, Wang, Wu, Wu, Wu, Xie, Yasunaga, You,
  Zaharia, Zhang, Zhang, Zhang, Zhang, Zheng, Zhou, and
  Liang]{bommasani2021opportunities}
R.~Bommasani, D.~A. Hudson, E.~Adeli, R.~Altman, S.~Arora, S.~von Arx, M.~S.
  Bernstein, J.~Bohg, A.~Bosselut, E.~Brunskill, E.~Brynjolfsson, S.~Buch,
  D.~Card, R.~Castellon, N.~Chatterji, A.~Chen, K.~Creel, J.~Q. Davis,
  D.~Demszky, C.~Donahue, M.~Doumbouya, E.~Durmus, S.~Ermon, J.~Etchemendy,
  K.~Ethayarajh, L.~Fei-Fei, C.~Finn, T.~Gale, L.~Gillespie, K.~Goel,
  N.~Goodman, S.~Grossman, N.~Guha, T.~Hashimoto, P.~Henderson, J.~Hewitt,
  D.~E. Ho, J.~Hong, K.~Hsu, J.~Huang, T.~Icard, S.~Jain, D.~Jurafsky,
  P.~Kalluri, S.~Karamcheti, G.~Keeling, F.~Khani, O.~Khattab, P.~W. Koh,
  M.~Krass, R.~Krishna, R.~Kuditipudi, A.~Kumar, F.~Ladhak, M.~Lee, T.~Lee,
  J.~Leskovec, I.~Levent, X.~L. Li, X.~Li, T.~Ma, A.~Malik, C.~D. Manning,
  S.~Mirchandani, E.~Mitchell, Z.~Munyikwa, S.~Nair, A.~Narayan, D.~Narayanan,
  B.~Newman, A.~Nie, J.~C. Niebles, H.~Nilforoshan, J.~Nyarko, G.~Ogut, L.~Orr,
  I.~Papadimitriou, J.~S. Park, C.~Piech, E.~Portelance, C.~Potts,
  A.~Raghunathan, R.~Reich, H.~Ren, F.~Rong, Y.~Roohani, C.~Ruiz, J.~Ryan,
  C.~Ré, D.~Sadigh, S.~Sagawa, K.~Santhanam, A.~Shih, K.~Srinivasan,
  A.~Tamkin, R.~Taori, A.~W. Thomas, F.~Tramèr, R.~E. Wang, W.~Wang, B.~Wu,
  J.~Wu, Y.~Wu, S.~M. Xie, M.~Yasunaga, J.~You, M.~Zaharia, M.~Zhang, T.~Zhang,
  X.~Zhang, Y.~Zhang, L.~Zheng, K.~Zhou, and P.~Liang.
\newblock On the opportunities and risks of foundation models.
\newblock \emph{arXiv preprint arXiv:2108.07258}, 2021.

\bibitem[Buzzega et~al.(2020)Buzzega, Boschini, Porrello, Abati, and
  Calderara]{buzzega2020dark}
P.~Buzzega, M.~Boschini, A.~Porrello, D.~Abati, and S.~Calderara.
\newblock Dark experience for general continual learning: a strong, simple
  baseline.
\newblock In \emph{Advances in Neural Information Processing Systems
  (NeurIPS)}, 2020.

\bibitem[Chaudhry et~al.(2019)Chaudhry, Ranzato, Rohrbach, and
  Elhoseiny]{chaudhry2018efficient}
A.~Chaudhry, M.~Ranzato, M.~Rohrbach, and M.~Elhoseiny.
\newblock Efficient lifelong learning with a-gem.
\newblock In \emph{Proceedings of the International Conference on Learning
  Representations (ICLR)}, 2019.

\bibitem[Christie et~al.(2018)Christie, Fendley, Wilson, and
  Mukherjee]{christie2018fmow}
G.~Christie, N.~Fendley, J.~Wilson, and R.~Mukherjee.
\newblock Functional map of the world.
\newblock In \emph{Computer Vision and Pattern Recognition (CVPR)}, 2018.

\bibitem[Deng et~al.(2009)Deng, Dong, Socher, Li, Li, and
  Fei-Fei]{deng2009imagenet}
J.~Deng, W.~Dong, R.~Socher, L.-J. Li, K.~Li, and L.~Fei-Fei.
\newblock Imagenet: A large-scale hierarchical image database.
\newblock In \emph{Proceedings of the IEEE International Conference on Computer
  Vision and Pattern Recognition (CVPR)}, 2009.

\bibitem[Fini et~al.(2022)Fini, da~Costa, Alameda-Pineda, Ricci, Alahari, and
  Mairal]{fini2022self}
E.~Fini, V.~G.~T. da~Costa, X.~Alameda-Pineda, E.~Ricci, K.~Alahari, and
  J.~Mairal.
\newblock Self-supervised models are continual learners.
\newblock In \emph{Proceedings of the IEEE International Conference on Computer
  Vision and Pattern Recognition (CVPR)}, 2022.

\bibitem[Houlsby et~al.(2019)Houlsby, Giurgiu, Jastrzebski, Morrone,
  De~Laroussilhe, Gesmundo, Attariyan, and Gelly]{adapterbert}
N.~Houlsby, A.~Giurgiu, S.~Jastrzebski, B.~Morrone, Q.~De~Laroussilhe,
  A.~Gesmundo, M.~Attariyan, and S.~Gelly.
\newblock Parameter-efficient transfer learning for {NLP}.
\newblock In K.~Chaudhuri and R.~Salakhutdinov, editors, \emph{Proceedings of
  the 36th International Conference on Machine Learning}, volume~97 of
  \emph{Proceedings of Machine Learning Research}, pages 2790--2799. PMLR,
  09--15 Jun 2019.
\newblock URL \url{https://proceedings.mlr.press/v97/houlsby19a.html}.

\bibitem[Jean et~al.(2016)Jean, Burke, Xie, Davis, Lobell, and
  Ermon]{jean2016combining}
N.~Jean, M.~Burke, M.~Xie, W.~M. Davis, D.~B. Lobell, and S.~Ermon.
\newblock Combining satellite imagery and machine learning to predict poverty.
\newblock \emph{Science}, 353, 2016.

\bibitem[Ke et~al.(2021)Ke, Xu, and Liu]{Ke2021AdaptingBF}
Z.~Ke, H.~Xu, and B.~Liu.
\newblock Adapting bert for continual learning of a sequence of aspect
  sentiment classification tasks.
\newblock In \emph{NAACL}, 2021.

\bibitem[Krizhevsky et~al.(2012)Krizhevsky, Sutskever, and
  Hinton]{krizhevsky2012imagenet}
A.~Krizhevsky, I.~Sutskever, and G.~E. Hinton.
\newblock Imagenet classification with deep convolutional neural networks.
\newblock In \emph{Advances in neural information processing systems}, pages
  1097--1105, 2012.

\bibitem[Kumar et~al.(2022)Kumar, Raghunathan, Jones, Ma, and Liang]{LPFT}
A.~Kumar, A.~Raghunathan, R.~Jones, T.~Ma, and P.~Liang.
\newblock Fine-tuning can distort pretrained features and underperform
  out-of-distribution.
\newblock In \emph{International Conference on Learning Representations}, 2022.
\newblock URL \url{https://openreview.net/forum?id=UYneFzXSJWh}.

\bibitem[Lopez-Paz and Ranzato(2017)]{lopez2017gradient}
D.~Lopez-Paz and M.~Ranzato.
\newblock Gradient episodic memory for continual learning.
\newblock In \emph{Advances in Neural Information Processing Systems
  (NeurIPS)}, 2017.

\bibitem[Madaan et~al.(2022)Madaan, Yoon, Li, Liu, and
  Hwang]{madaan2022representational}
D.~Madaan, J.~Yoon, Y.~Li, Y.~Liu, and S.~J. Hwang.
\newblock Representational continuity for unsupervised continual learning.
\newblock In \emph{International Conference on Learning Representations}, 2022.
\newblock URL \url{https://openreview.net/forum?id=9Hrka5PA7LW}.

\bibitem[Rao et~al.(2019)Rao, Visin, Rusu, Teh, Pascanu, and
  Hadsell]{rao2019continual}
D.~Rao, F.~Visin, A.~A. Rusu, Y.~W. Teh, R.~Pascanu, and R.~Hadsell.
\newblock Continual unsupervised representation learning.
\newblock In \emph{Advances in Neural Information Processing Systems
  (NeurIPS)}, 2019.

\bibitem[Rusu et~al.(2016)Rusu, Rabinowitz, Desjardins, Soyer, Kirkpatrick,
  Kavukcuoglu, Pascanu, and Hadsell]{rusu2016progressive}
A.~A. Rusu, N.~C. Rabinowitz, G.~Desjardins, H.~Soyer, J.~Kirkpatrick,
  K.~Kavukcuoglu, R.~Pascanu, and R.~Hadsell.
\newblock Progressive neural networks.
\newblock \emph{arXiv preprint arXiv:1606.04671}, 2016.

\bibitem[Sinha et~al.(2020)Sinha, Song, Garg, and Ermon]{sinha2020experience}
S.~Sinha, J.~Song, A.~Garg, and S.~Ermon.
\newblock Experience replay with likelihood-free importance weights.
\newblock \emph{arXiv preprint arXiv:2006.13169}, 2020.

\bibitem[Wu et~al.(2018)Wu, Xiong, Yu, and Lin]{wu18knn}
Z.~Wu, Y.~Xiong, S.~X. Yu, and D.~Lin.
\newblock Unsupervised feature learning via non-parametric instance
  discrimination.
\newblock In \emph{Proceedings of the IEEE International Conference on Computer
  Vision and Pattern Recognition (CVPR)}, 2018.

\bibitem[Yoon et~al.(2018)Yoon, Yang, Lee, and Hwang]{YoonJ2018iclr}
J.~Yoon, E.~Yang, J.~Lee, and S.~J. Hwang.
\newblock Lifelong learning with dynamically expandable networks.
\newblock In \emph{Proceedings of the International Conference on Learning
  Representations (ICLR)}, 2018.

\bibitem[Yoon et~al.(2020)Yoon, Kim, Yang, and Hwang]{yoon2020apd}
J.~Yoon, S.~Kim, E.~Yang, and S.~J. Hwang.
\newblock Scalable and order-robust continual learning with additive parameter
  decomposition.
\newblock In \emph{Proceedings of the International Conference on Learning
  Representations (ICLR)}, 2020.

\bibitem[Zenke et~al.(2017)Zenke, Poole, and Ganguli]{zenke17si}
F.~Zenke, B.~Poole, and S.~Ganguli.
\newblock Continual learning through synaptic intelligence.
\newblock In \emph{Proceedings of the International Conference on Machine
  Learning (ICML)}, 2017.

\end{thebibliography}
\bibliographystyle{abbrvnat}
}

\newpage
\appendix
\begin{center}
\begin{huge}
\textbf{Appendix}
\end{huge}
\end{center}
\section{Full Fewshot results}
\label{app:app-1}

\begin{table}[h!]
\caption{We test each method’s utility on 10\% few-shot data under our various probes’ evaluation scheme. Nearly all methods obtain a better probe when we apply LPFT at test time, but finetuning and LPFT yield high-performing probes that beats (CIFAR100) or nears DER and SI on CIFAR10. The ranking of the probes change significantly across the columns. DER and SI is dominant under the traditional KNN and linear probe scheme, but fades in comparison on LPFT probe scheme. Using additional data at test time helps “simple” methods defeat their more involved counterparts, without incurring additional cost like memory from replay buffer (DER) or information accumulation (SI).}
\begin{subtable}{0.99\columnwidth}\centering
\caption{}
\label{fewshot-1}
\resizebox{\linewidth}{!}{
\begin{tabular}{lllll}
\hline
     & \multicolumn{2}{c}{KNN probe}                                                            & \multicolumn{2}{c}{LP probe}                                                              \\ \hline
     & \textsc{C10}              & \textsc{C100}              & \textsc{C10}               & \textsc{C100}              \\ \hline
\textsc{FT}   & 5. 84.51 (+-0.58)                          & 5. 70.76 (+-0.64)                           & 5. 90.69 (+-0.22)                           & 4. 85.90 (+-0.20)                           \\
\textsc{UCL}  & 3. 88.53 (+0.05)                           & 4. 76.14 (+0.06)                            & 4. 91.63 (+-0.22)                           & 5. 84.78 (+-0.16)                              \\
\textsc{LPFT} & 4. 88.45 (+-0.15)                          & 3. 79.45 (+-0.70)                           & 3. 91.70 (+-0.14)                           & \textbf{1. 88.60 (+0.42)}  \\ \hline
\textsc{SI}   & \textbf{1. 91.90 (+0.08)} & 2. 79.83 (+-0.01)                           & \textbf{1. 93.44 (+-0.21)} & \textbf{2. 88.38 (+-0.36)} \\
\textsc{DER}  & 2. 88.98 (+-0.35)                          & \textbf{1. 80.58 (+-0.11)} & 2. 93.01 (+-0.15)                           & \textbf{3. 88.34 (+-0.11)} \\ \hline
\end{tabular}}
\end{subtable}
\begin{subtable}{0.99\columnwidth}\centering
\caption{}
\label{fewshot-2}
\resizebox{\linewidth}{!}{
\begin{tabular}{lllll}
\hline
                               & \multicolumn{2}{c}{LPFT probe}                                                            & \multicolumn{2}{c}{Best probe}                                                            \\ \hline
                               & \textsc{C10}               & \textsc{C100}              & \textsc{C10}               & \textsc{C100}              \\ \hline
\textsc{FT}   & 2. 94.93 (+0.10)                            & 2. 91.99 (+-0.16)                           & 2. 94.93 (+-0.10)                           & 2. 91.99 (+-0.16)                           \\
\textsc{UCL}  & 5. 89.15 (+0.90)                            & 5. 84.97 (+-0.25)                           & 5. 91.63 (+-0.22)                           & 5. 84.78 (+-0.16)                           \\
\textsc{LPFT} & 3. 94.84 (+-0.11)                           & \textbf{1. 92.63 (+-0.17)} & 3. 94.84 (+-0.11)                           & \textbf{1. 92.63 (+-0.17)} \\ \hline
\textsc{SI}   & 4. 93.36 (+-0.08)                           & 4. 88.73 (+-0.24)                           & 4. 93.44 (+-0.21)                           & 4. 88.73 (+-0.24)                           \\
\textsc{DER}  & \textbf{1. 95.21 (+-0.11)} & 3. 91.31 (+-0.11)                           & \textbf{1. 95.21 (+-0.11)} & 3. 91.31 (+-0.11)                           \\ \hline
\end{tabular}}
\end{subtable}

\label{tab:fewshot-probes-eval-full}
\end{table}

\begin{figure}[h!]
    \centering
    \includegraphics[width=.8\columnwidth]{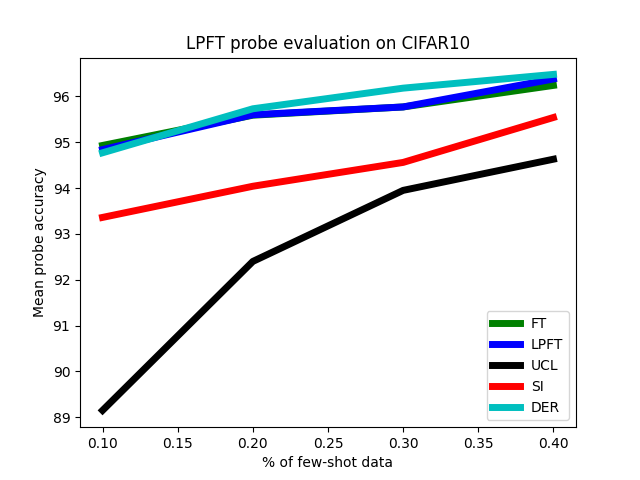}
    \caption{To better evaluate the representations, we retrain the probes for few-shot evaluation using a small percentage of training data. We plot the percent of data along x-axis, and observe the increasing accuracy trend on the y-axis. We note that LPFT and FT are both competitive with DER, and do not suffer the drawback of unsupervised CL in the low-data regime.}
    \label{fig:lpft-probe-eval}
\end{figure}

\newpage
\section{Vision and NLP Architecture}
\label{app:app-2}
We use a Resnet18 (parameters) for our C10, C100, Tiny experiments and Densenet121 for FMOW. For the NLP sequential datasets, we use the same pretrained BERT as \cite{Ke2021AdaptingBF}.

In the first case, only the final classification head is specific to a task, and so we train only those weights before finetuning the whole network for every new task. Adapter modules were introduced for large pretrained models in NLP by \cite{adapterbert}, and \cite{Ke2021AdaptingBF} built B-CL to yield a compact and extensible model. It is designed to add a few trainable parameters (task mask) to each encoder layer per new task, in a way the model can be extended while a) contributing shared knowledge to the learnt representations, and b) mitigating forgetting by protecting the neurons in the task-specific module using a task-specific mask. In the second case, we train only the task mask for every new task, then finetune the whole network after. Following their notations, we tune the task-specific embedding $e_l^{(t)}$ described in section 4.3.

We also find a new checkpoint strategy per task to produce better results for both \naive and LPFT. Previously, the model checkpoint after training the full number of epochs per task (10 epochs for the 2 SemEval tasks, 30 epochs for other tasks) is passed on to the next task. Instead, we select the task checkpoint which achieves the highest average accuracy over all tasks up to and including the current task. We also find training 30 epochs for SemEval datasets improves both \naive and LPFT. The final \naive accuracy, 89.51\% is 1.22\% higher than \cite{Ke2021AdaptingBF}'s reported accuracy. For the SemEval task datasets, using LPFT is not sufficient for convergence, so we use \naive on those 2 tasks in our LPFT runs. \\

\noindent Our code is available at \href{https://github.com/AnanyaKumar/UCL_michael}{link1} and \href{https://github.com/shiningsunnyday/UCL}{link2}.

\end{document}